# A Memetic Algorithm Based on Breakout Local Search for the Generalized Travelling Salesman Problem

Mehdi El Krari [*], Belaïd Ahiod

Faculty of Science, Mohammed V University in Rabat
Rabat, Morocco
LRIT, associated unit to CNRST (URAC 29)

abbreviated title: A Memetic BLS for the GTSP

**Abstract**

The Travelling Salesman Problem (TSP) is one of the most popular Combinatorial Optimization Problem. It is well solicited for the large variety of applications that it can solve, but also for its difficulty to find optimal solutions. One of the variants of the TSP is the Generalized TSP (GTSP), where the TSP is considered as a special case which makes the GTSP harder to solve. We propose in this paper a new memetic algorithm based on the well-known Breakout Local Search (BLS) metaheuristic to provide good solutions for GTSP instances. Our approach is competitive compared to other recent memetic algorithms proposed for the GTSP and gives at the same time some improvements to BLS to reduce its runtime.

## 1 Introduction

The Generalized Travelling Salesman Problem (GTSP) is an extension of the well-known Travelling Salesman Problem (TSP), they are both Combinatorial Optimization Problems (COP). The problem has numerous applications such as postal routing, airport selection, sequencing computer files and many others (Laporte, Asef-Vaziri, and Sriskandarajah 1996), the work made by Ilhan (2017) is a concrete example of real-life applications of such problems. Several approaches and studies have been considered since GTSP was presented by Henry-Labordere (1969) and Srivastava et al. (1969). Fischetti, Salazar Gonzalez, and Toth (1997) introduced branch-and-cut which is an exact algorithm. Such algorithms are important but very greedy in runtime. Lin-Kernighan, one of the most successful heuristic for the TSP, has been adapted for the GTSP by Karapetyan and Gutin

[*] Corresponding author: M. El Krari
Email: mehdi@elkrari.com

(2011). The only metaheuristics that competed with Lin-Kernighan in the TSP are those combine genetic and local searches algorithms, called memetic algorithms (Hart, Krasnogor, and Smith 2004). Last and best studies for the GTSP focused on memetic with different local search algorithms and genetic approaches.

We introduce in our paper a new memetic algorithm based on the Breakout Local Search (BLS) as the main method for running local searches. Iterated Local Search (ILS) (Lourenço, Martin, and Stutzle 2001) is known to suffer from lack of effectiveness in escaping some local optima, BLS follows the basic scheme of this framework and improves it by combining the features of other efficient methods during perturbation process. BLS find a local optimum with a local search procedure, then run the most appropriate perturbations to jump from one neighbourhood to another in the search space. Those perturbations are done with a variable degree of diversification by determining dynamically how many jumps to perform and making an adaptive selection from several types of dedicated movements. BLS perturbation strategy is based on both the history and state of search.

BLS was developed by Benlic and Hao in 2012. Since then, it has been used for solving some COP, such as the Quadratic Assignment Problem (Benlic and Hao 2013b), Maximum Clique Problems (Benlic and Hao 2013a), Max-Cut problem (Benlic and Hao 2013c) and Minimum Sum Coloring Problem (Benlic and Hao 2012), and has given very good results. BLS was also used in other studies of by other researchers (Ghandi and Masehian 2015; Fu and Hao 2014). This paper provides a memetic algorithm based on BLS metaheuristic for the GTSP, whose performance will be evaluated by solving 39 benchmark instances of the GTSPLIB (Reinelt 1991).

We will talk in the next Section about the GTSP by defining the problem and presenting some previous works. We introduce later in Section 3 our memetic approach where we devote an important part to the BLS metaheuristic which is the main contribution. Experimental tests and comparisons are exposed in Section 4 and finally a conclusion in Section 5.

## 2 The Generalized Traveling Salesman Problem

### 2.1 Problem formulation

The GTSP can be considered as a weighted complete graph $G = (V, E, c)$ where V = $(v_1, v_2, ..., v_n)$ represents the set of vertices/cities, E is the set of all edges/paths ($v_i$ , $v_j$ ) for *i, j = 1, 2, ...n* and $c$ is the cost function E $\rightarrow \mathbb{N}^+$ ( or $\mathbb{R}^+$) giving the weight for each edge which is the distance between two nodes. While the TSP consists of visiting all the given nodes V, the GTSP nodes are distributed over a set of clusters C = $(C_1, C_2, ...C_m)$ such that and the intersection of any pair of clusters is an empty set.

A GTSP solution can be formulated as a sub-graph $S = (N, D) \in G$. $N = \{n_1, n_2, ...n_m\}$ is the set of nodes where each cluster $C_k$ has one node $n_t$ for k, t = 1, 2, ...m. $D \in E$ is the subset of edges composed by N and forming a cyclic tour which total distance is calculated by the formula (1) below:

$$W(T) = c(p_m, p_1) + \sum_{i=1}^{m-1} c(p_i, p_{i+1}) \qquad (1)$$

## 2.2 Previous works

First solutions provided for the GTSP were methods based on dynamic programming (Srivastava et al. 1969; Henry-Labordere 1969). Laporte, Mercure, and Nobert (1987) proposed an exact method using integer programming (Branch and Bound) techniques. Later, Fischetti, Salazar Gonzalez, and Toth (1997) made a solution based on Branch and Cut which provide optimal solutions for any instance with up to 442 cities and 89 clusters.

Several researchers (Ben-Arieh et al. 2003; Laporte and Semet 1999; Noon and Bean 1993) suggested transformations of a GTSP instance into a TSP one as there are efficient heuristics and exact algorithms for the TSP. This idea is very limited since we must have exact solutions from obtained TSP instance, otherwise, it may lead to an infeasible GTSP solution.

A variety of tour construction heuristics inspired from the TSP have been adapted on the GTSP, Noon and Bean (1991) proposed a nearest-neighbour heuristic, while Fischetti, Salazar Gonzalez, and Toth (1997) presented other adaptations such as the farthest, nearest and cheapest insertion.

Several local search heuristics were studied in (Karapetyan and Gutin 2011, 2012; Pourhassan and Neumann 2015; Smith and Imeson 2017) and some others. Many researchers worked also on Genetic Algorithm(s) (GA) (Bontoux, Artigues, and Feillet 2010; Snyder and Daskin 2006; Silberholz and Golden 2007) while others made hybrid algorithms with local search heuristics and Genetic Algorithms yielding memetic algorithms.

Memetic algorithms (Hart, Krasnogor, and Smith 2004) are known for their competitiveness since they hybrid between local searches which favour exploitation and genetic algorithms which are good for diversification. It's known in the TSP that the only metaheuristics that can compete with the famous Lin-Kernighan are memetic algorithms. This fact lets many researchers focus on memetic algorithms for the GTSP (Gutin, Karapetyan, and Krasnogor 2008; Bontoux, Artigues, and Feillet 2010; Gutin and Karapetyan 2010) and motivates us to work on it.

# 3 Memetic Breakout Local Search

## 3.1 The BLS framework

BLS is a metaheuristic based on the Iterated Local Search (ILS) framework: it consists of discovering the search space by moving from a neighbourhood to a new one via perturbation(s) once the algorithm meets a local optimum. BLS provides an interesting strategy which establishes perturbations whose moves and number of jumps are elected dynamically on each iteration.

From an initial permutation $\pi_0$ (usually produced by a construction method), BLS land in a local optimum via a local search procedure. It browses the entire neighbourhood to find the best improving permutation to alter the current local optimum. The best move found during the iteration is saved in a history matrix H which will be solicited later during the algorithm.

In addition to the history matrix, BLS perturbations are determined by two other parameters: (i) L decides how many jumps to perform in each perturbation. It is incremented if the previous one didn't help to escape the current neighbourhood, otherwise downgraded to a minimal value $L_0$. (ii) $\omega$ counts the number of the consecutive non-improving best solution found so far. It is incremented by one if the best permutation is not updated, else it will be reset to zero. $\omega$ has a maximal value T, if reached then BLS run a strong perturbation which resides on performing a perturbation with an important number of jump $L_{max}$, then $\omega$ is reset to zero. Algorithm 1 below describes the overall approach of the BLS framework.

Every local search ends in a local optimum and needs diversification to explore a new neighbourhood. Perturbations are an important step in the ILS framework (and thus for BLS too) since they allow to jump to a new area in the search space. The distance between this area and the current local optimum depends on the nature of the perturbation. BLS introduces a new well adapted adaptive perturbations defined by the nodes/cities to be involved in the movement and how many perturbations to perform before to start a new local search.

**Algorithm 1**: Breakout Local Search Framework

**Require**: $\pi$: initial solution, **Desc$_{max}$**: maximal descents to perform, **L$_0$**: initial number of jumps, **T**: maximal consecutive visited local optima without any improvement, **L$_{max}$**: number of jumps when reaching T

**Ensure**: A solution $\pi_{best}$

1.     c ← objectiveValue($\pi$)
2.     $\pi_{best}$ ← $\pi$    /* $\pi_{best}$ saves best solution found */
3.     $c_{best}$ ← c    /* $c_{best}$ saves best objective value */
4.     $\omega$ ← 0    /* $\omega$ gives the number of consecutive non-improving best solution found so far */
5.     L ← L$_0$    /* L saves number of jumps to perform, set to it's minimal value L$_0$ */
6.     $c_p$ ← c    /* $c_p$ saves objective value of last descent */
7.     Desc ← 0    /* Desc saves current number of descents */
8.     Iter ← 0    /*global iteration counter*/
9. **while** Desc ⟨ *Desc$_{max}$* **do**
10.       **while** ∃ 2optMove(x,y) such that (c+delta2Opt($\pi$,x,y)⟨c) **do**
11.           $\pi$ ← $\pi$ ⊕ 2OptMove(x,y)    /* perform the best improving move*/
12.           c ← c + delta2Opt($\pi$,x,y)    /* cost variation of $\pi$ with (x,y) move*/
13.           Update H with iteration number when edges move was last performed
14.           Iter ← Iter+1
15.       **end while**
16.       **if** c ⟨ $c_{best}$ **then**
17.           Update $\pi_{best}$ and $c_{best}$
18.           $\omega$ ← 0    /* reset counter for consecutive non-improving local optima */
19.       **else**
20.           $\omega$ ← $\omega$+1
21.       **end if**
22.       **if** $\omega$ ⟩ T **then**    /*strong perturbation with L$_{max}$ moves */
23.           $\pi$ ← Perturbation($\pi$, L$_{max,}$ H, Iter, $\omega$, M)
24.           $\omega$ ← 0
25.       **else if** c = $c_p$ **then**    /* search returned the previous local optimum */
26.           L ← L+1    /*increment moves if the previous move didn't help to improve */
27.       **else**    /*Search escaped from the previous local optimum, reinitialize indicator */
28.           L ← L$_0$
29.       **end if**
30.       $c_p$ ← c    /* update the objective value of the previous local optimum */
31.       **if** strong perturbation not performed **then**
32.           $\pi$ ← Perturbation($\pi$, L, H, Iter, $\omega$, M)    /* see algorithm 2 */
33.       **end if**
34. **end while**
35.     **return** $\pi_{best}$

### 3.1.1 Principle of adaptive perturbations

*Main idea*

Perturbations in the ILS framework must be well chosen. Indeed, perturbations with low intensity allow to explore gradually the search space, but maybe insufficient when the local optimum is attractive which may imply a stagnation in the current neighbourhood. While strong perturbations ensure to escape the neighbourhood but may seem to be a new random start. BLS proposes perturbations that adapt on the search state, the number of jumps and movements depend on how long the search is trapped in the current neighbourhood.

Instead of completing arbitrarily moves in each iteration, BLS come up with three different perturbations: (i) directed, (ii) recency-based and (iii) random. Each one is called in specific circumstances following a probability and generates a set M of moves that will be solicited during the perturbation with the constraint that the resulting tour must be feasible.

*The three types of perturbation moves*

**The directed perturbation** is designed to select movement candidates resulting to a new solution with minimal degradation. These movements should not belong to a tabu list (Glover 1989, 1990) (which length is γ) unless one of them can produce a new solution which fitness is better than the best found so far. Candidates for the directed perturbation are set by the set A:

$$A = \{swapMove(u,v) | min\{deltaSwap(\pi, u, v)\},$$
$$(H_{uv} + \gamma) < Iter \vee (deltaSwap(\pi, u, v) + c) < c_{best}, u \neq v\} \quad (2)$$

The **recency-based perturbation** is, as indicated by his name, established from the history of movements which is saved in H. The selected moves are the oldest ones or those who have never been solicited if there are. This candidates' set is defined by the set B where

$$B = \{swapMove(u,v) | min\{H_{uv}\}, u \neq v\} \quad (3)$$

The **random perturbation** is the third and last one in the BLS adaptive perturbation. Candidates (C) are chosen randomly without criteria...

$$C = \{swapMove(u,v) | u \neq v\} \quad (4)$$

BLS selects for each jump one of the three perturbations described above. The selection process is done pseudo-randomly following a probability calculated by ω and T. Directed perturbation have a high probability to be selected when the current neighbourhood is recently discovered (ω is then small). This probability becomes lower when the local optimum seems to be more attractive ($\omega$ begins to tend towards T), recency-based and random perturbations are then privileged to offer a stronger diversification.

Using the directed perturbation is done with a probability P which has a threshold $P_0$ (Eq. (5)). While the recency-based and the random perturbations are decided respectively by $(1-P) \times Q$ and $(1-P) \times (1-Q)$ where Q is a constant from [0,1].

$$P = \begin{cases} e^{-\omega/T}, & if(e^{-\omega/T} > P_0) \\ P_0, & otherwise \end{cases} \quad (5)$$

From a local optimum ($\pi$), BLS perform a sequence of L jumps, each one selects its move from a set of candidates built by one the three formulas (2-4). Current solution and moves History are updated after each jump and if ever BLS meet a solution that improves the best one found so far, $\pi_{best}$ is updated and $\omega$ is reset to zero.

---

**Algorithm 2: Perturbation($\pi$, L, H, Iter, $\omega$, M)**

---

**Require:** $\pi$: current solution, which is a local optimality, **L**: Number of jumps, **H**: Matrix of moves historic, **Iter**: Global iteration counter, **$\omega$**: Number of consecutive non-improving local optima, **M**: Set of candidate moves
**Ensure:** A perturbed solution $\pi$
1.     **for** i from 1 to L **do**
2.         take a pair(x,y) $\in$ M
3.         **update** $\pi$, c, and H
4.         Iter $\leftarrow$ Iter + 1
5.         **if** c $<$ $c_{best}$ **then**
6.             $\pi_{best} \leftarrow \pi$
7.             $c_{best} \leftarrow c$
8.             $\omega \leftarrow 0$
9.         **end if**
10.     **end for**

---

In each iteration of the perturbation process, BLS builds the set M of candidates following one of the three perturbations. The directed and recency-based perturbations need to browse all the history matrix which cost $O(n^2)$. Building the set M "L times" during only one perturbation impacts the runtime especially for large instances and make BLS very slow.

Instead of looking through all the matrix H, we propose in our adaptation for the GTSP to pick randomly *N* moves and keep those who will satisfy the selected perturbation. This improvement will decrease the complexity of each jump from $O(n^2)$ to $O(n)$ and will highly contribute reducing the BLS runtime.

## 3.2 The memetic BLS

We propose in our work a memetic algorithm (MA) by combining the studied BLS (in subsection 3.1) with the benefits of genetic algorithm (GA) (Davis 1991).

A GA follows the natural mechanism of selection. It consists of constructing a population of candidate solutions, called chromosomes, which will be the first generation. The population converges gradually from a generation of chromosomes to new ones to reach a final solution having the best objective value from all met candidates.

The generations evolve at each iteration of the GA heuristic. Several procedures are run each time on the chromosomes to enhance the population by finding better individuals that can improve the overall fitness and become the best solution found so far. We expose in this subsection all the components that compose a MA framework.

### 3.2.1 Initialization

M/2 solutions are generated to constitute the population, where M is the number of clusters. Each solution is created with the semi-random construction heuristic. It consists of generating a random permutation of clusters and select the best node in each cluster. This can be done using the Cluster Optimization heuristic (Fischetti, Salazar Gonzalez, and Toth 1997).

### 3.2.2 Improvements

Improvements in MA algorithms are usually made with one or many local search procedures. Each solution in the population will be improved by BLS in order to get a population with good quality solutions.

### 3.2.3 Crossover

The crossover operator produces a new solution(s) for the population by coupling two existing solutions ($S_1$ and $S_2$) belonging to the previous generation and selected following a strategy. Uniform crossover (UC) (Syswerda 1989) is used in our memetic approach, it produces two children from $S_1$ and $S_2$ which are picked from the population by a tournament selection (Miller, Goldberg, and others 1995), each one is selected among three solutions picked randomly.

### 3.2.4 Mutation

The mutation is a genetic operator that alters some solutions in the population in order to reduce the probability of convergence by maintaining a diversified population. The mutation is performed on a solution without any criteria and may naturally guide to the worst solution.

We use in our memetic approach the double bridge move (Martin, Otto, and Felten 1991) as a mutation operator in order to minimize chances of coming back to the same solution after the improvement step.

### 3.2.5 Termination condition

We stop producing new generations once we found the best-known fitness for the concerned instance *I*. Else, the maximum number of generations is equal to the number of clusters in *I*. The output solution will be the best solution in the last generation.

## 4 Experimental results

### 4.1 Experimental protocol

Memetic BLS is implemented in Java 1.7, and run on a Pentium Dual-Core CPU T4400 with 2.20 GHz and 2.8 GB of memory. 39 instances from the TSPLIB benchmark (and converted to GTSP instances by the standard clustering procedure (Fischetti et al., 1997)) are solved whose sizes are from 10 clusters and 48 nodes to 89 clusters and 442 nodes, each one is run 20 times.

### 4.2 Computational results and comparisons

Memetic BLS is compared during our tests with another memetic approach by (Bontoux, Artigues, and Feillet 2010) and a random-key genetic algorithm proposed by (Snyder and Daskin 2006). Below are listed the performance measures used for evaluating our Memetic BLS and comparing it with the branch-and-cut runtime (Fischetti, Salazar Gonzalez, and Toth 1997) and the two other approaches.

(i) The average deviation of obtained solutions from the best known solution, denoted $dev : (dev = 100 \times (f_{avg} - best)/best[\%])$ where $f_{avg}$ is the fitness average of the 20 solutions obtained from 20 runs of Memetic BLS, and *best* is the best known fitness.

(ii) the CPU time in seconds.

Table 1: Comparative results between Memetic BLS and other evolutionary heuristics

| Instance | Nodes | Clusters | Best | B&C Time | MBLS | | Bontoux | | Snyder | |
|---|---|---|---|---|---|---|---|---|---|---|
| | | | | | dev | CPU (s) | dev | CPU (s) | dev | CPU (s) |
| 10att48 | 48 | 10 | 5,394 | 2.1 | - | - | **0.00** | 0.57 | **0.00** | 0.00 |
| 10gr48 | 48 | 10 | 1,834 | 1.9 | - | - | **0.00** | 0.97 | **0.00** | 0.50 |
| 10hk48 | 48 | 10 | 6,386 | 3.8 | - | - | **0.00** | 0.58 | **0.00** | 0.20 |
| 11eil51 | 51 | 11 | 174 | 2.9 | **0.00** | 0.05 | **0.00** | 0.84 | **0.00** | 0.10 |
| 11berlin52 | 52 | 11 | 4,040 | - | **0.00** | 0.03 | - | - | - | - |
| 12brazil58 | 58 | 12 | 15,332 | 3.0 | - | - | **0.00** | 0.85 | **0.00** | 0.30 |
| 14st70 | 70 | 14 | 316 | 7.3 | **0.00** | 0.09 | **0.00** | 1.02 | **0.00** | 0.20 |
| 16eil76 | 76 | 16 | 209 | 9.4 | **0.00** | 0.42 | **0.00** | 1.18 | **0.00** | 0.20 |
| 16pr76 | 76 | 16 | 64,925 | 12.9 | **0.00** | 0.64 | **0.00** | 1.27 | **0.00** | 0.20 |
| 20gr96 | 96 | 20 | 29,440 | 19.4 | **0.00** | 1.32 | - | - | - | - |
| 20kroA100 | 100 | 20 | 9,711 | 18.4 | **0.00** | 0.57 | **0.00** | 1.98 | **0.00** | 0.40 |
| 20kroB100 | 100 | 20 | 10,328 | 22.2 | **0.00** | 0.68 | **0.00** | 2.01 | **0.00** | 0.40 |
| 20kroC100 | 100 | 20 | 9,554 | 14.4 | **0.00** | 0.60 | **0.00** | 1.84 | **0.00** | 0.30 |
| 20kroD100 | 100 | 20 | 9,450 | 14.3 | **0.00** | 0.56 | **0.00** | 2.93 | **0.00** | 0.40 |
| 20kroE100 | 100 | 20 | 9,523 | 13.0 | **0.00** | 1.21 | **0.00** | 1.87 | **0.00** | 0.80 |
| 20rat99 | 99 | 20 | 497 | 51.5 | **0.00** | 3.13 | **0.00** | 3.52 | **0.00** | 0.70 |
| 20rd100 | 100 | 20 | 3,650 | 16.6 | **0.00** | 2.12 | **0.00** | 2.93 | **0.00** | 0.30 |
| 21eil101 | 101 | 21 | 249 | 25.6 | **0.00** | 1.14 | **0.00** | 2.07 | **0.00** | 0.20 |
| 21lin105 | 105 | 21 | 8,213 | 16.4 | **0.00** | 0.64 | **0.00** | 3.25 | **0.00** | 0.30 |
| 22pr107 | 107 | 22 | 27,898 | 7.4 | **0.00** | 0.68 | **0.00** | 4.67 | **0.00** | 0.40 |
| 24gr120 | 120 | 24 | 2,769 | 41.9 | - | - | **0.00** | 2.3 | **0.00** | 0.50 |
| 25pr124 | 124 | 25 | 36,605 | 25.9 | **0.00** | 2.65 | **0.00** | 2.89 | **0.00** | 0.60 |
| 26ch130 | 130 | 26 | 2,828 | - | **0.00** | 5.74 | - | - | - | - |
| 26bier127 | 127 | 26 | 72,418 | 23.6 | **0.00** | 1.33 | **0.00** | 3.02 | **0.00** | 0.50 |
| 28gr137 | 137 | 28 | 36,417 | - | **0.00** | 6.09 | - | - | - | - |
| 28pr136 | 136 | 28 | 42,570 | 43.0 | **0.00** | 6.34 | **0.00** | 4.17 | **0.00** | 0.50 |
| 29pr144 | 144 | 29 | 45,886 | 8.2 | **0.00** | 1.75 | **0.00** | 5.38 | **0.00** | 0.30 |
| 30ch150 | 150 | 30 | 2,750 | - | **0.00** | 9.49 | - | - | - | - |
| 30kroA150 | 150 | 30 | 11,018 | 100.3 | **0.00** | 15.77 | **0.00** | 5.27 | **0.00** | 1.30 |
| 30kroB150 | 150 | 30 | 12,196 | 60.6 | **0.00** | 10.21 | **0.00** | 4.61 | **0.00** | 1.00 |

| Instance | Nodes | Clusters | Best | B&C Time | MBLS | | Bontoux | | Snyder | |
|---|---|---|---|---|---|---|---|---|---|---|
| | | | | | dev | CPU (s) | dev | CPU (s) | dev | CPU (s) |
| 31pr152 | 152 | 31 | 51,576 | 94.8 | **0.00** | 4.09 | 0.00 | 4.45 | 0.00 | 1.50 |
| 32u159 | 159 | 32 | 22,664 | 146.4 | **0.00** | 8.46 | 0.00 | 5.53 | 0.00 | 0.60 |
| 39rat195 | 195 | 39 | 854 | 245.9 | **0.00** | 9.55 | 0.00 | 10.42 | 0.00 | 0.70 |
| 40d198 | 198 | 40 | 10,557 | 763.1 | **0.00** | 34.00 | 0.00 | 8.72 | 0.00 | 1.20 |
| 40kroA200 | 200 | 40 | 13,406 | 187.4 | **0.00** | 45.00 | 0.00 | 6.74 | 0.00 | 2.70 |
| 40kroB200 | 200 | 40 | 13,111 | 268.5 | **0.00** | 71.00 | 0.00 | 8.78 | 0.00 | 1.40 |
| 41gr202 | 202 | 41 | 23,301 | 1,022 | **0.00** | 81.00 | - | - | - | - |
| 45ts225 | 225 | 45 | 68,340 | 37,875 | **0.00** | 141.00 | 0.04 | 35.31 | 0.00 | 2.40 |
| 46gr229 | 229 | 46 | 71,972 | 1,187 | **0.00** | 56.00 | - | - | - | - |
| 46pr226 | 226 | 46 | 64,007 | 106.9 | - | - | 0.00 | 6.92 | 0.00 | 1.00 |
| 53gil262 | 262 | 53 | 1,013 | 6,624 | **0.09** | 346.00 | 0.14 | 25.12 | 0.79 | 1.90 |
| 53pr264 | 264 | 53 | 29,549 | 337.0 | - | - | 0.00 | 16.64 | 0.00 | 1.30 |
| 56a280 | 280 | 56 | 1,079 | - | **0.12** | 3861.00 | - | - | - | - |
| 60pr299 | 299 | 60 | 22,615 | 812.8 | - | - | 0.00 | 20.19 | 0.00 | 6.10 |
| 64lin318 | 318 | 64 | 20,765 | 1,671 | - | - | 0.00 | 24.89 | 0.00 | 3.50 |
| 80rd400 | 400 | 80 | 6,361 | 7,021 | - | - | **0.42** | 38.33 | 1.37 | 3.50 |
| 84fl417 | 417 | 84 | 9,651 | 16,719 | 0.17 | 4733.00 | **0.00** | 21.9 | 0.07 | 2.40 |
| 88pr439 | 439 | 88 | 60,099 | 5,422 | **0.00** | 2733.00 | 0.00 | 56.46 | 0.23 | 9.10 |
| 89pcb442 | 442 | 89 | 21,657 | 58,770 | 1.28 | 2437.00 | **0.19** | 76.64 | 1.31 | 10.10 |
| **Average** | | | | | 0.04 | 375.21 | 0.02 | 10.46 | 0.09 | 1.46 |

Experimental results done on the Memetic BLS are quite satisfying. Our approach succeeds to reach the optimal solution 35 times from the 39 studied instances, while the average deviation (for all instances) is 0.04% which let our method competitive regarding the others using in the comparison. On the other hand, memetic BLS seems to be greedy in runtime which not strange for BLS (Benlic and Hao 2013b). Recall that BLS has been improved in this work and without this improvement, runtime could be much higher.

# 5 Conclusion

We presented in this paper a new memetic approach for solving the GTSP where the BLS metaheuristic is used in the improvement stage. BLS is a performing metaheuristic that proved its performances in many COPs. Results given in all related works (including this one) are very convincing especially in terms of fitness and deviation from the optimal solution.

The exposed results in the previous Section demonstrate the competitiveness of the memetic BLS by finding optimal solutions in most of the studied instances. The comparative study done consolidates the good performances of our work. The metaheuristic complexity still restrictive because of some heavy components which can be considered as a drawback, depending on decision makers. Thus, we proposed an improvement to the perturbation component by reducing the complexity from $O(n^2)$ to $O(n)$.